\documentclass{article}
\pdfoutput=1
\usepackage[a4paper, margin=1in]{geometry}

\usepackage{times}
\usepackage{graphicx} 
\usepackage{subfigure}

\usepackage{algorithm}
\usepackage{algorithmic}

\usepackage{authblk}

\usepackage{hyperref}
\usepackage{url}
\usepackage{amsmath, amssymb, amsthm}
\usepackage{verbatim}
\usepackage{stmaryrd} 
\usepackage{xspace}
\usepackage{setspace}
\usepackage{subfigure}
\usepackage{color}

\makeatletter
\DeclareRobustCommand\onedot{\futurelet\@let@token\@onedot}
\def\@onedot{\ifx\@let@token.\else.\null\fi\xspace}
\def\iid{{i.i.d}\onedot}
\def\eg{{e.g}\onedot} 
\def\ie{{i.e}\onedot}

\DeclareMathOperator*{\E}{\operatorname{\mathbb{E}}}

\newcommand{\X}{\mathcal{X}}
\newcommand{\Y}{\mathcal{Y}}

\newcommand{\er}{\operatorname{er}}
\newcommand{\eer}{\widehat{\er}}

\newcommand{\disc}{\operatorname{disc}}

\renewcommand{\H}{\mathcal{H}} 
\newcommand{\supp}{\operatorname{supp}} 

\newcommand{\bZ}{\mathbf{Z}}
\newcommand{\bS}{\mathbf{S}}
\newcommand{\bh}{\mathbf{h}}
\newcommand{\bpi}{\boldsymbol{\pi}}
\newcommand{\balpha}{\boldsymbol{\alpha}}

\newtheorem{theorem}{Theorem}
\newtheorem{lemma}{Lemma}

\newtheorem{proposition}{Proposition}

\title{Multi-Task Learning with Labeled and Unlabeled Tasks} 

\author{Anastasia Pentina}
\author{Christoph H. Lampert \\ \texttt{\{apentina,chl\}@ist.ac.at}}
\affil{IST Austria (Institute of Science and Technology Austria)}
\date{}

\begin{document} 
\maketitle

\begin{abstract} 
In multi-task learning, a learner is given a collection of prediction 
tasks and needs to solve all of them. 
In contrast to previous work, which required that annotated training 
data is available for all tasks, we consider a new setting, in which 
for some tasks, potentially most of them, only unlabeled training 
data is provided.
Consequently, to solve all tasks, information must be transferred 
between tasks with labels and tasks without labels. 
Focusing on an instance-based transfer method we analyze 
two variants of this setting: when the set of labeled tasks 
is fixed, and when it can be \emph{actively} selected 
by the learner. 
We state and prove a generalization bound that covers both scenarios 
and derive from it an algorithm for making the choice of labeled 
tasks (in the active case) and for transferring information between 
the tasks in a principled way.
We also illustrate the effectiveness of the algorithm by experiments 
on synthetic and real data.
\end{abstract} 

\section{Introduction}
In the \emph{multi-task learning} setting~\cite{Caruana} 
a learner is given a collection of prediction tasks 
that all need to be solved. %
The hope is that the overall prediction quality can be improved by processing the tasks jointly and sharing information between them.
Indeed, theoretical as well as experimental studies have shown that information transfer can reduce the amount of annotated examples per task needed to achieve good performance under various assumptions on how the learning tasks are related.

All existing multi-task learning approaches have in common, however, that they need at least some labeled training data for every task of interest. 
In this paper, we study a new and more challenging setting, in which for a subset of the tasks (typically the large majority) only unlabeled data is available.
In practice, it is highly desirable to be able to handle this situation for problems with a very large number of tasks, such as sentiment analysis for market studies: for different products different attributes matter and, thus, each product should be have 
its own predictor and forms its own learning task. At the same time annotating data for each such task is prohibitive, especially when new products are constantly added to the market.
Another example are prediction problems, for which the \emph{fixed cost} of obtaining any labels for a task can be high, even when the \emph{variable cost} per label are reasonable. 
This is a well-known issue when using crowd sourcing for data annotation: recruiting and training annotators first imposes a large overhead, and only afterwards many labels can be obtained within a short time and at a low cost. 

A distinctive feature of the setting we study is that it requires two types of information transfer: between the labeled tasks and from labeled to unlabeled ones. 
While the first type is common in multi-task learning, none of the existing multi-task methods is able to handle the second type.
In contrast, information transfer from labeled to unlabeled tasks is commonly studied in domain adaptation research, where, however, transfer of the first type is typically not considered.
Thus, the setting of \emph{multi-task learning with labeled and unlabeled tasks} can be seen as a blend of traditional multi-task learning and domain adaptation.

In this work we focus on a transfer method that learns a predictor for every task of interest by minimizing a task-specific convex combination of training errors on the labeled tasks~\cite{citeulike:3449264,DBLP:conf/colt/MansourMR09}.
We choose this method because it allows us to capture both types of information transfer -- between the labeled tasks and from labeled to unlabeled ones -- in a unified fashion.
Clearly, the success of this approach depends on the choice of the weights in the convex combinations.
Moreover, one can expect it also to depend on the subset of labeled tasks as well, because some subsets of tasks might be more informative and representative than the others.
%
%
%
This suggests that it will be beneficial if the labeled subset is not arbitrary but if it can be chosen in a data-dependent way. 
We refer to this learning scenario, where initially every task is represented only by a set of unlabeled examples and the learner can choose for which tasks to request some labels, as \emph{active task selection}.

Our main result is a generalization bound that quantifies both of the aforementioned effects: it relates the total multi-task error to quantities that depend on the subset of labeled tasks and on the task-specific weights used for information transfer. 
%
%
Using the computable quantities in the bound as an objective function and minimizing it numerically, we obtain a principled algorithm for selecting which tasks to have labeled (in the active task selection scenario) and for choosing task-specific weights and predictors for all tasks, labeled as well as unlabeled. 
We highlight the practical usefulness of the derived method by experiments on synthetic and real data. 

The success of any information transfer approach, regardless whether it is applied in the multi-task or the domain adaptation scenario, depends on the relatedness between tasks of interest.
Indeed, one cannot expect to benefit from information transfer between the labeled tasks or to be able to obtain solutions of reasonable quality for the unlabeled ones if the given tasks are completely unrelated.
An advantage of the method we propose is that from the associated generalization bound we can read off explicitly under which conditions the algorithm can be expected to succeed.
In particular, it suggests that the proposed method is likely to succeed if the given set of tasks satisfies the following assumption of \emph{task smoothness}: if two tasks are similar in their marginal distributions, then their optimal prediction functions are also likely to be similar.
A more formal definition will be given in Section~\ref{sec:interpretation}. 
The task smoothness assumption resembles the classical \emph{smoothness assumption} of semi-supervised learning~\cite{chapelle2006semi}.
It can be expected to hold in many real-world settings with a large number of tasks, for example in the aforementioned case of sentiment analysis: if two products are described using similar words, these words would likely have similar connotation for both products.  
Note, also, that a similar assumption appears implicitly in~\cite{NIPS2011_1205}. 

\subsection{Related Work}
Most existing multi-task learning methods work in the fully supervised setting and aim at improving the overall prediction quality by sharing information between the tasks. 
For this they employ different types of transfer: \emph{instance-transfer} methods re-use training samples from different tasks~\cite{CrammerNIPS2012}, \emph{parameter-transfer} methods assume that the predictors for all tasks are similar to each other in some norm and exploit this fact through specific regularizers~\cite{Evgeniou}, \emph{representation-transfer} approaches assume that the predictors for all tasks share a common (low-dimensional) representation that can be learned from the data~\cite{evgeniou2007multi,Argyriou}. 
Follow-up works extended and generalized these concepts, \eg by learning the relatedness of tasks~\cite{journals/jmlr/SahaRDV11,kang2011learning} or sharing only between subgroups of tasks~\cite{xue2007multi,Kumar,barzilai2015convex}. 
However, all of the above methods require at least some labeled data for each task.

To our knowledge, the only existing multi-task method that can be applied in the considered setting where for some tasks only unlabeled data is available is~\cite{ECCV12_Khosla}.
Motivated by the problem of dataset bias, this method relies on the assumption that different tasks are minor modifications (\ie biased versions) of the same, true prediction problem.
Similarly to~\cite{Evgeniou}, it uses specific regularizers and trains predictors for all tasks jointly as small perturbations of a common predictor, which corresponds to the hypothetical unbiased task and can potentially be applied to unseen problems.
Thus, applied in the considered setting, this method provides one predictor for all unlabeled tasks and treats the labeled ones as slight variations of them. 

Information transfer from labeled to unlabeled tasks is the question typically studied in domain adaptation research.
In fact, if the set of labeled tasks is fixed, any domain adaptation technique might be used to obtain solutions for unlabeled tasks, in particular those based on source reweighting~\cite{shimodaira2000improving}, representation learning~\cite{pan2011domain,glorot2011domain}, or semi-supervised transfer~\cite{xing2007bridged}.
However, by design all domain adaptation methods aim at finding the best predictor on a \emph{single} target task given a \emph{fixed} set of source tasks.
Therefore none of them can readily be applied in the active task selection setting, where the learner needs to \emph{select} the labeled tasks that would lead to good performance across \emph{all tasks}.

A second related setting is zero-shot learning~\cite{larochelle2008zero,lampert-tpami2013,palatucci2009zero}, where contextual, usually semantic, information is used to solve a learning task for which no training data is available. 
The situation we are interested in is more specific than this, though, as we assume that unlabeled data of the tasks is available, not context in an arbitrary form. As we will show, this allows us to derive formal performance guarantees that zero-shot learning methods typically lack. 

The active task selection scenario is directly related to the question of identifying a representative set of source tasks in domain adaptation, a question that has previously been raised in the context of sentiment analysis~\cite{Blitzer07biographies}.
It also shares some features with \emph{active learning}, where the learner is given a set of unlabeled samples and can choose a subset to obtain labels for.
A fundamental difference is, however, that in active learning the learner needs to find a single prediction function for all labeled and unlabeled data while in the multi-task setting each task, including unlabeled ones, potentially requires its own predictor.
%
%
In the multi-task or zero-shot setting, active learning has so far not found widespread use. 
Exemplary works in this direction are~\cite{reichart2008multi,saha2010active,gavves2015active}, which, however, use active learning on the level of training examples, not tasks. 
The idea of choosing tasks was used in active curriculum selection~\cite{AAAI136463,pentina2015curriculum}, where the learner can influence the order in which tasks are processed.
However these methods nevertheless require annotated examples for all tasks of interest.
%

%

\section{MTL with Labeled and Unlabeled Tasks}
\label{sec:theory}
In the multi-task setting the learner observes a collection of prediction tasks and its goal is to learn all of them. 
Formally, we assume that there is a set of $T$ tasks $\{\langle D_1,f_1\rangle,\dots, \langle D_T, f_T\rangle\}$, where each task $t$ is defined by a marginal distribution $D_t$ over the input space $\X$ and a deterministic labeling function $f_t:\X\rightarrow\Y$.
The goal of the learner is to find $T$ predictors $h_1,\dots,h_T$ in a hypothesis set $\H\subset\{h:\X\rightarrow\Y\}$ that would minimize the average expected risk:
\begin{align}
\label{eq:average_exp_risk}
\er(h_1,\dots,h_T)&= \frac{1}{T}\sum_{t=1}^T\er_t(h_t),\\
\text{where}\;\;
\er_t(h_t)&=\E_{x\sim D_t}\ell(h_t(x),f_t(x)).
\notag
\end{align} 

In this work we concentrate on the case of binary classification tasks, $\Y=\{-1,1\}$, and $0/1$-loss, $\ell(y_1,y_2)=0$
if $y_1=y_2$, and $\ell(y_1,y_2)=1$ otherwise.

In the fully-supervised setting the learner is given a training set of annotated examples for every task of interest.
In contrast, we consider the scenario where every task $t$ is represented by a set $S_t=\{x^t_1,\dots,x^t_n\}$ of $n$ unlabeled examples sampled \iid according to the marginal distribution $D_t$.
For a subset of $k$ tasks $\{i_1,\dots,i_k\}$, which are either predefined or, in the active scenario, can be selected based on the unlabeled data, the learner is given labels for a random subset $\overline{S}_{i_j}\subset S_{i_j}$ of $m$ points.

To obtain a predictor for any task, labeled or unlabeled, we consider a method that minimizes a convex combination of training errors of the labeled tasks.
This choice allows us to capture, in a unified fashion, both types of information transfer -- between the labeled tasks and from labeled to unlabeled ones.
Formally, for a set of tasks $I=\{i_1,\dots,i_k\}\subset\{1,\dots,T\}$ we define:
\begin{equation}
\Lambda^I\!=\!\left\{\alpha\in[0,1]^T: \sum_{i=1}^T\alpha_i=1; \: \supp\alpha\subseteq I\right\}
\end{equation}
for $\supp\alpha=\{i\in\{1,\dots,T\}: \alpha_i\neq 0\}$.
Given a weight vector $\alpha\in\Lambda^I$, the $\alpha$-weighted empirical error of a hypothesis $h\in\H$ is defined as follows:
\begin{align}
\eer_\alpha(h)&=\sum_{i\in I}\alpha_i\eer_i(h),\label{eq:weighted-emp-error}\\
\text{where}\;\;
\eer_i(h)&=\frac{1}{m}\sum_{(x,y)\in \overline{S}_i}\ell(h(x),y).
\end{align}
%
In order to obtain a solution for any task $t$ the learner minimizes $\eer_{\alpha^t}(h)$ for some $\alpha^t\in\Lambda^I$, where $I$ is the set of labeled tasks, potentially in combination with some regularization. 

The success of this approach depends on the subset $I$ of tasks that are labeled and on the weights $\alpha^1,\dots,\alpha^T$.
The following theorem quantifies both of these effects and will later be used to chose $\alpha^1,\dots,\alpha^T$ and potentially $I$ in a principled way.

\begin{theorem}
Let $d$ be the VC dimension of the hypothesis set $\H$, $k$ be the number of labeled tasks, $S_1,\dots,S_T$ be $T$ sets of size $n$ each, where $S_i\stackrel{\iid}{\sim}D_i$, and $\overline{S}_1,\dots,\overline{S}_T$ be their random subsets of size $m$ each, for which labels would be provided if the corresponding task is selected as labeled.
Then for any $\delta>0$ with probability at least $1-\delta$ over $S_1,\dots,S_T$ and $\overline{S}_1,\dots,\overline{S}_T$ uniformly for all choices of labeled tasks $I=\{i_1,\dots,i_k\}$ and weights $\alpha^1,\dots,\alpha^T\in\Lambda^I$, provided that they are fully determined by the unlabeled data only, and for all possible choices of $h_1,\dots,h_T\in \H$ the following inequality holds:
\begin{align}
\label{eq:main}
&\frac{1}{T}\!\sum_{t=1}^T\!\er_{t}(h_t)\!\leq\!\frac{1}{T}\!\sum_{t=1}^T\!\eer_{\alpha^t}(h_t)\!+\!\frac{1}{T}\!\sum_{t=1}^T\!\sum_{i\in I}\!\alpha^t_{i}\disc(S_t,S_{i})+\frac{A}{T}\!\|\alpha\|_{2,1}+\!\frac{B}{T}\!\|\alpha\|_{1,2}+\!C\!+\!D\!+\!\frac{1}{T}\!\sum_{t=1}^T\sum_{i\in I}\alpha^t_{i}\lambda_{ti},
\end{align}
where
\begin{align*}
\disc(S_t,S_i)&=\max_{h,h'\in\H}|\eer_{S_t}(h,h')-\eer_{S_i}(h,h')|
\end{align*}
with $\eer_{S_i}(h,h')=\frac{1}{n}\sum_{j=1}^n\ell(h(x^i_j),h'(x^i_j))$
is the empirical discrepancy between unlabeled samples $S_t$ and $S_i$, and
\begin{align*}
\lambda_{ij}&=\min_{h\in\H}(\er_i(h)+\er_j(h))\\
\|\alpha\|_{2,1} &= \sum_{t=1}^T\sqrt{\sum_{i\in I}(\alpha^t_{i})^2},\;\;
\|\alpha\|_{1,2} = \sqrt{\sum_{i\in I}\left(\sum_{t=1}^T\alpha^t_{i}\right)^2},\\
A &= \sqrt{\frac{2d\log(ekm/d)}{m}},\;\;
B = \sqrt{\frac{\log(4/\delta)}{2m}}\\
C &= \sqrt{\frac{8(\log T+d\log(enT/d))}{n}}+\sqrt{\frac{2}{n}\log\frac{4}{\delta}},\\
D &=2\sqrt{\frac{2d\log(2n)+2\log(T) + \log(4/\delta)}{n}}.
\end{align*}
\label{thm:main}
\end{theorem} 

\textbf{Proof Sketch} (the full proof can be found in Appendix~\ref{app:proof}).
By Theorem 2 in~\cite{Ben-David:2010}, for any two tasks $t$ and $i$ the following inequality holds for every $h\in\H$:
\begin{equation}
\er_t(h)\leq\er_i(h)+\disc(D_t,D_i)+\lambda_{ti}.
\end{equation}
Thus, we obtain the following bound on the average expected error over all tasks in terms  of the error on the labeled tasks:
\begin{align}
\label{eq:multi-source-non-probabilistic}
\frac{1}{T}\!\sum_{t=1}^T\er_t(h_t)\!&\leq\!\frac{1}{T}\!\sum_{t=1}^T\!\er_{\alpha^t}(h_t)+\frac{1}{T}\!\sum_{t=1}^T\sum_{i\in I}\!\alpha^t_{i}\disc(D_t,D_{i})
+\frac{1}{T}\sum_{t=1}^T\sum_{i\in I}\alpha^t_{i}\lambda_{ti},
\\\text{with}\;\;
\er_{\alpha^t}(h_t)&=\sum_{i\in I}\alpha^t_i\E_{x\sim D_i}\ell(h_t(x),f_i(x)),
\;\;\text{and}
\\
\er_{D_i}(h,h')&=\E_{x\sim D_i}\ell(h(x),h'(x)),\;\;\text{and}
\\
\label{eq:discrepancy}
\disc(D_t,D_i)&\!= \!\max_{h,h'\in\H}\!\!|\er_{D_t}(h,h')\!-\!\er_{D_i}(h,h')|\!
\end{align}
is the discrepancy between two distributions~\cite{Kifer,DBLP:conf/colt/MansourMR09,Ben-David:2010}.
In order to prove the statement of the theorem we need to relate the $\alpha$-weighted expected errors and discrepancies between the marginal distributions in~\eqref{eq:multi-source-non-probabilistic} to their empirical estimates.

The proof consists of three steps.
First, we show that, conditioned on the unlabeled data, $\frac{1}{T}\sum_{t=1}^T\tilde{\er}_{\alpha^t}$ can be upper bounded in terms of  $\frac{1}{T}\sum_{t=1}^T\eer_{\alpha^t}$,
where:
\begin{equation*}
\tilde{\er}_{\alpha}(h)=\sum_{i\in I}\alpha_i\tilde{\er}_{i}(h)=\sum_{i\in I}\frac{\alpha_i}{n}\sum_{j=1}^n\ell(h(x^i_j),f_i(x^i_j)).
\end{equation*}
This quantity can be interpreted as a training error if the learner would receive the labels for all the samples for the chosen tasks $I$.
Note that in case of $m=n$ this step is not needed and we can avoid the corresponding complexity terms.
In the second step we relate  the average $\alpha$-weighted expected errors to $\frac{1}{T}\sum_{t=1}^T\tilde{\er}_{\alpha^t}$.
In the third step we conclude the proof by bounding the pairwise discrepancies in terms of their empirical estimates.

\textbf{Step 1.} 
Fix the unlabeled sets $S_1,\dots,S_T$.
They fully determine the choice of labeled tasks $I$ and the weights $\alpha^1,\dots,\alpha^T$. 
Therefore, conditioned on the unlabeled data, these quantities can be considered constant and the bound has to hold uniformly only with respect to $h_1,\dots,h_T$.

%
In order to simplify the notation we assume that $I=\{1,\dots,k\}$ and define:
\begin{equation}
\Phi(\overline{S}_1,\dots,\overline{S}_k) =\sup_{h_1,\dots,h_T} \frac{1}{T}\sum_{t=1}^T\tilde{\er}_{\alpha^t}(h_t)-\eer_{\alpha^t}(h_t).
\end{equation}
Note that one could analyze this quantity using standard techniques from Rademacher analysis, if the labeled examples were sampled from the unlabeled sets \iid, \ie with replacement.
However, since we assume that for every $i$ $\overline{S}_i$ is a subset of $S_i$, \ie the labeled examples are sampled randomly without replacement, there are dependencies between the labeled examples.
Therefore we utilize techniques from the literature on transductive learning~\cite{DBLP:conf/colt/El-YanivP07} instead.
We first apply Doob's construction to $\Phi$ in order to obtain a martingale sequence and then use McDiarmid's inequality for martingales~\cite{mcdiarmid89}.
As a result we obtain that with probability at least $1-\delta/4$ over sampling labeled examples: 
\begin{equation}
\Phi\leq\E_{\overline{S}_1,\dots,\overline{S}_k}\Phi + \frac{1}{T}\sqrt{\sum_{i=1}^k\left(\sum_{t=1}^T\alpha^t_i\right)^2} \sqrt{\frac{\log(4/\delta)}{2m}}.
\label{eq:step1.1}
\end{equation}
Now we need to upper bound $\E\Phi$. 
Using results from~\cite{TolBlaKlo14} and~\cite{Hoeffding:1963} we observe that:
\begin{equation}
\E_{\overline{S}_1,\dots,\overline{S}_k}\Phi(\overline{S}_1,\dots,\overline{S}_k)\leq\E_{\tilde{S}_1,\dots,\tilde{S}_k}\Phi(\tilde{S}_1,\dots,\tilde{S}_k),
\end{equation}
where $\tilde{S}_i$ is a set of $m$ points sampled from $S_i$ \iid with replacement (in contrast to sampling without replacement corresponding to $\overline{S}_i$).
This means that we can upper bound the expectation of $\Phi$ over samples with dependencies by the expectation over independent samples. 
By doing so, applying the symmetrization trick, and introducing Rademacher random variables, we obtain that:
\begin{equation}
\E_{\overline{S}_1,\dots,\overline{S}_k}\Phi\leq\frac{1}{T}\sum_{t=1}^T\sqrt{\sum_{i=1}^k(\alpha^t_i)^2}\cdot\sqrt{\frac{2d\log(ekm/d)}{m}}.
\label{eq:step1.2}
\end{equation}
A combination of~\eqref{eq:step1.1} and~\eqref{eq:step1.2} shows that (conditioned on the unlabeled data) with probability at least $1-\delta/4$ over sampling labeled examples uniformly for all choices of $h_1,\dots,h_T$ the following holds:
\begin{equation}
\frac{1}{T}\!\sum_{t=1}^T\!\tilde{\er}_{\alpha^t}(h_t)\leq\frac{1}{T}\!\sum_{t=1}^T\eer_{\alpha^t}(h_t)+1\frac{A}{T}\|\alpha\|_{2,1}+\frac{B}{T}\|\alpha\|_{1,2}.
\label{eq:step1}
\end{equation}

\textbf{Step 2.} Now we relate $\frac{1}{T}\sum_{t=1}^T\tilde{\er}_{\alpha^t}$ to $\frac{1}{T}\sum_{t=1}^T\er_{\alpha^t}$.

The choice of the tasks to label, $I$, the corresponding weights, $\alpha$, and the predictors, $h$, all depend on the unlabeled data.
Therefore, we aim for a bound that is uniform in all three parameters.
We define: 
\begin{gather*}
\Psi(S_1,\dots,S_T) =\\
\sup_{I}\sup_{\alpha^1,\dots,\alpha^T\in\Lambda^I}\sup_{h_1,\dots,h_T}
\frac{1}{T}\sum_{t=1}^T\sum_{i=1}^T\alpha^t_{i}(\er_{i}(h_t)-\tilde{\er}_{i}(h_t)).
\end{gather*}
%
The main instrument that we use here is a refined version of McDiarmid's inequality, which is due to~\cite{Maurer:2006}.
It allows us to use the standard Rademacher analysis, while taking into account the internal structure of the weights $\alpha^1,\dots,\alpha^T$.
As a result we obtain that with probability at least $1-\delta/4$ simultaneously for all choices of tasks to be labeled, $I$, weights $\alpha^1,\dots,\alpha^T\in\Lambda^I$ and hypotheses $h_1,\dots,h_T$:
\begin{align}
\frac{1}{T}\sum_{t=1}^T\er_{\alpha^t}(h_t)\leq\frac{1}{T}\sum_{t=1}^T\tilde{\er}_{\alpha^t}(h_t)+C.
\label{eq:step2}
\end{align}
%

\textbf{Step 3.} To conclude the proof we bound the pairwise discrepancies in terms of their finite sample estimates.
According to Lemma 1 in~\cite{Ben-David:2010} for any pair of tasks $i,j$ and any  $\delta>0$ with probability at least $1-\delta$:
\begin{equation*}
\disc(D_i,D_j)\leq\disc(S_i,S_j)+2\sqrt{\frac{2d\log(2n)+\log(2/\delta)}{n}}.
\end{equation*}
We apply this result to every pair of tasks and combine the results using the uniform bound argument.
This yields the remaining two terms on the right hand side: the weighted average of the sample-based discrepancies and the constant $D$.
By combining the result with~\eqref{eq:step1} and~\eqref{eq:step2} we obtain the statement of the theorem.
\qed

\section{Explanation and Interpretation}
\label{sec:interpretation}

The left-hand side of inequality~\eqref{eq:main} is the average expected error over all $T$ tasks, the quantity of interest that the learner would like to minimize but cannot directly compute.
It is upper-bounded by the sum of two complexity terms and five task-dependent terms: weighted training errors on the labeled tasks, weighted averages of the distances to the labeled tasks in terms of the empirical discrepancies, two mixed norms of the weights $\alpha$ and a weighted average of $\lambda$-s.

The complexity terms $C$ and $D$ behave as $O(\sqrt{d\log(nT)/n})$ and converge to zero when the number of unlabeled examples per task, $n$, tends to infinity.
In contrast, $\frac{A}{T}\|\alpha\|_{2,1}+\frac{B}{T}\|\alpha\|_{1,2}$ in the worst case of $\|\alpha\|_{2,1}=\|\alpha\|_{1,2}=T$ behaves as $O(\sqrt{d\log(km)/m})$ and converges to zero when the number of labeled examples per labeled task, $m$, tends to infinity.
In order for these terms to be balanced, \ie for the uncertainty coming from the estimation of discrepancy to not dominate the uncertainty from the estimation of the $\alpha$-weighted risks, the number of unlabeled examples per task $n$ should be significantly (for $k\ll T$) larger than $m$.
However, this is not a strong limitation under the common assumption that obtaining enough unlabeled examples is significantly cheaper than annotated ones.

\begin{figure*}[th]\centering
\fbox{\includegraphics[width=.28\textwidth]{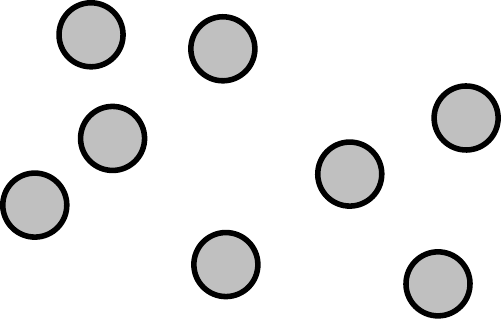}}\quad
\fbox{\includegraphics[width=.28\textwidth]{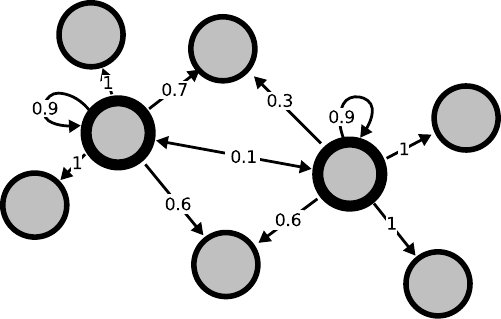}}\quad
\fbox{\includegraphics[width=.28\textwidth]{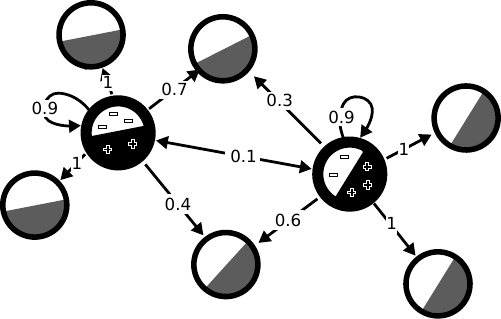}}
\caption{Schematic illustration of \emph{active task selection}.
Left: eight unlabeled tasks need to be solved. Center: the subset of tasks to be labeled and between-task weights
are determined by minimizing \eqref{eq:objective}. Right: annotated data for labeled tasks is obtained, and 
prediction functions (black vs.\ white) for each task are learned using the learned weighted combinations. Sharing can occur between labeled tasks.
}
\label{fig:multi-task}
\end{figure*}

The remaining terms on the right-hand side of~\eqref{eq:main} depend on the set of labeled tasks $I$, the tasks-specific weights $\alpha$-s and hypotheses $h$-s.
Thus, by minimizing them with respect to these quantities one can expect to obtain values for them that are beneficial for solving all tasks of interest based on the given data.
For the theorem to hold, the set of labeled tasks and the weights may not depend on the labels.  
The part of the bound that can be estimated based on the unlabeled data only, and therefore to select $I$ (in the active scenario) and $\alpha^1,\dots,\alpha^T$ is:
\begin{equation}
\frac{1}{T}\sum_{t=1}^T\sum_{i\in I}\alpha^t_{i}\disc(S_t,S_{i})+\frac{A}{T}\|\alpha\|_{2,1}+\frac{B}{T}\|\alpha\|_{1,2}.
\label{eq:objective}
\end{equation}
The first term in~\eqref{eq:objective} is the average weighted distance from every task to the labeled ones, as measured by the discrepancy between the corresponding unlabeled training samples.
This term suggests that for every task $t$ the largest weight, \ie the highest impact in terms of information transfer, should be put on a labeled task $i$ that has a similar marginal distribution. 
Note that the employed "similarity", which is captured by the discrepancy, directly depends on the considered hypothesis class and loss function and, thus, is tailored to a particular setting of interest. 
%
%
At the same time, the mixed-norm terms $\|\alpha\|_{1,2}$ and $\|\alpha\|_{2,1}$ prevent the learner from putting all weight on the single closest labeled task and can be seen as some form of regularization.
In particular, they encourage information transfer also between the labeled tasks, since minimizing just the first term in~\eqref{eq:objective} for every labeled tasks $i\in I$ would result in all weight to be 
put on task $i$ itself and nothing on other tasks, because by definition $\disc(S_i,S_i)=0$. 

The first mixed-norm term, $\|\alpha\|_{2,1}$ influences every $\alpha^t$ independently and encourages the learner to use data from multiple labeled tasks for adaptation.
Thus, it captures the intuition that sharing from multiple labeled tasks should improve the performance.
In contrast, $\|\alpha\|_{1,2}$ connects the weights for all tasks.
This term suggests to label tasks that all would be equally useful, thus preventing spending resources on tasks that would be informative for only a few of the remaining ones.
Also, it prevents the learner from having super-influential labeled tasks that share with too many others.
Such cases would be very unstable in the worst case scenario: mistakes on such tasks would propagate and have a major effect on the overall performance.

The effect of the mixed-norm terms can also be seen through the lens of the convergence rates.
Indeed, as already mentioned above, in the case of every $\alpha^t$ having only one non-zero component, $\|\alpha\|_{2,1}$ and $\|\alpha\|_{1,2}$ are equal to $T$ and thus the convergence rate\footnote{$\tilde{O}(\cdot)$ is an analog of $O(\cdot)$ that hides logarithmic factors} is $\tilde{O}(\sqrt{1/m})$.
However, in the opposite extreme, if every $\alpha^t$ weights all the labeled tasks equally, \ie $\alpha^t_i=1/k$ for all $t\in\{1,\dots,T\}$ and $i\in I$, then $\|\alpha\|_{2,1}=\|\alpha\|_{1,2}=\frac{T}{\sqrt{k}}$ and the convergence rate improves to  $\tilde{O}(\sqrt{1/km})$, which is the best one can expect from having a total of $km$ labeled examples.

The only term on the right-hand side of~\eqref{eq:main} that depends on the hypotheses $h_1,\dots,h_T$ and can be used to make a favorable choice is the weighted training error on the labeled tasks.
Thus, the generalization bound of Theorem~\ref{thm:main} suggest the following algorithm (Figure~\ref{fig:multi-task}): 
\\
\\
\fbox{
\parbox{0.96\linewidth}{\raggedright
\textbf{Algorithm 1.}

1. estimate pairwise discrepancies between the tasks based on the unlabeled data\\
2. choose the tasks $I$ to be labeled (in the active case)  and the weights  $\alpha^1,\dots,\alpha^T$ by minimizing~\eqref{eq:objective}\\ 
3. receive labels for the labeled tasks $I$\\
4. for every task $t$ train a classifier by minimizing~\eqref{eq:weighted-emp-error}  using the obtained weights $\alpha^t$.
}}%

Note, that this procedure is justified by Theorem~\ref{thm:main}:
all choices are done in agreement with the conditions of the theorem and, because the inequality~\eqref{eq:main} holds uniformly for all eligible choices of labeled tasks, weights and predictors, the guarantees also hold for the resulting solution. 

Algorithm 1 is guaranteed to perform well, if the solution it finds leads to a low value of the right-hand side of~\eqref{eq:main}.
By construction, it minimizes all data-dependent terms in~\eqref{eq:main}, except for one quantity that cannot be estimated from the available data:
\begin{equation}
\frac{1}{T}\sum_{t=1}^T\sum_{i\in I}\alpha^t_i\lambda_{ti}.
\label{eq:weighted-lambdas}
\end{equation}   
While discrepancy captures the similarity between marginal distributions, the $\lambda$-terms reflect the similarity between labeling functions: for every pair of task, $t$, and labeled task, $i\in I$, the corresponding value $\lambda_{ti}$ is small if there exists a hypothesis that performs well on both tasks.
Thus, Algorithm 1 can be expected to perform well,  if for any two given tasks $t$ and $i$ that are close to each other in terms of discrepancy (and thus in the minimization of~\eqref{eq:objective} the corresponding $\alpha^t_i$ is large), there exists a hypothesis that performs well on both of them (\ie the corresponding $\lambda_{ti}$ is small).
We refer to this property of the set of learning tasks as \emph{task smoothness}. 

%
%
%
%

Training predictors for every task of interest using data from all labeled tasks improves the statistical guarantees of the learner.
However, it results in empirical risk  minimization on up to $km$ samples for $T$ different weighted combinations.
Since we are most interested in the situation when $T$ is large, one might be interested in way to reduce the amount of necessary computation.
One way to do so is to drop the mixed-norm terms from the objective function~\eqref{eq:objective}, in which case it reduces to
\begin{equation}
\frac{1}{T}\sum_{t=1}^T\sum_{i\in I}\alpha^t_{i}\disc(S_t,S_{i}).
\label{eq:objective2}
\end{equation}
This expression is linear in $\alpha$ and thus minimizing it for a fixed set $I$ will lead to assigning each task to a single labeled task that is closest to it in terms of empirical discrepancy. Each labeled task will be assigned to itself.
Consequently, the learner must train only $k$ predictors, one for each labeled task, using only its $m$ samples.
The expression~\eqref{eq:objective2} can be seen as the $k$-medoids clustering objective with tasks 
corresponding to points in the space with (semi-)metric defined by empirical discrepancy and 
labeled tasks correspond to the centers of the clusters.
Thus, this method reduces to $k$-medoids clustering, resembling the suggestion of~\cite{Blitzer07biographies}.
Note that, nevertheless, the conditions of Theorem~\ref{thm:main} are fulfilled, and thus its guarantees 
will hold for the obtained solution.
The guarantees will be more pessimistic, however, than those from Algorithm 1, as the minimization
ignores parts of the bound~\eqref{eq:main} and will not use the potentially beneficial transfer between labeled tasks. 

\section{Experiments}\label{sec:experiments}

\begin{figure*}[t]\centering\label{fig:experiments}
\subfigure[Synthetic data, complete transfer]{\quad\includegraphics[width=.4\textwidth]{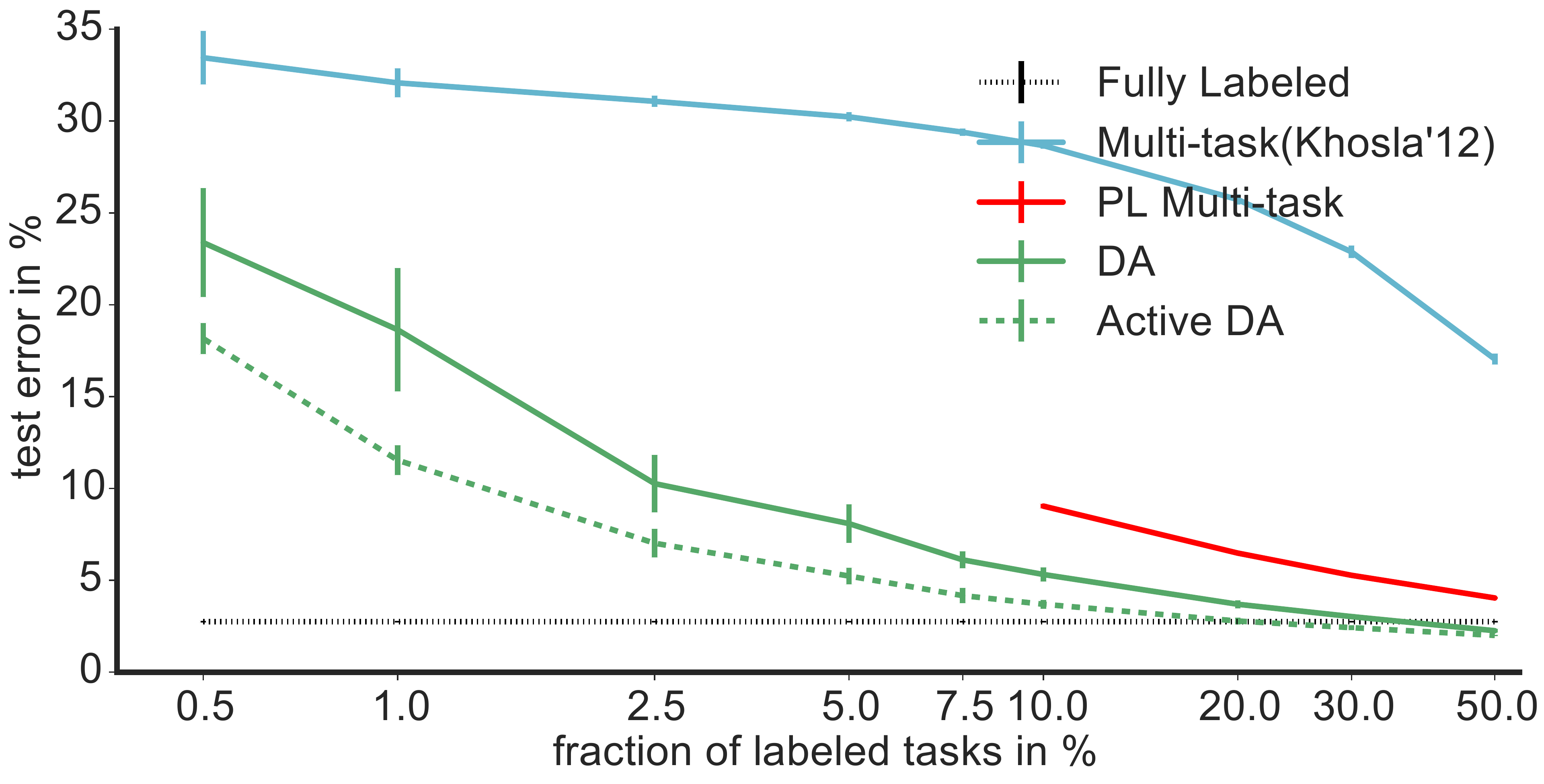}\quad}\label{fig:synthMS}
\subfigure[Product reviews, complete transfer]{\quad\includegraphics[width=.4\textwidth]{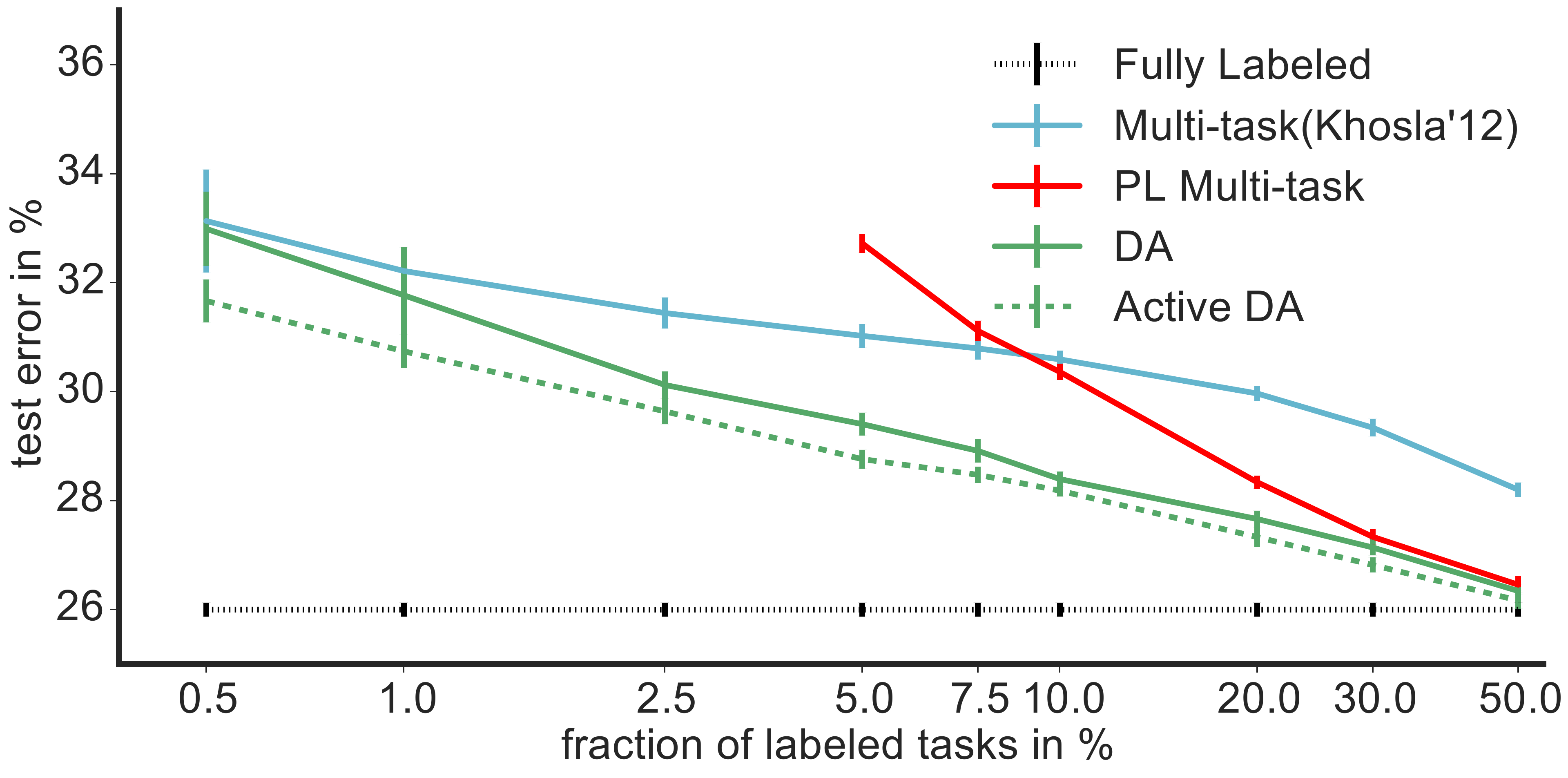}\quad}\label{fig:imagenetMS}
\subfigure[Synthetic data, single-source transfer]{\quad\includegraphics[width=.4\textwidth]{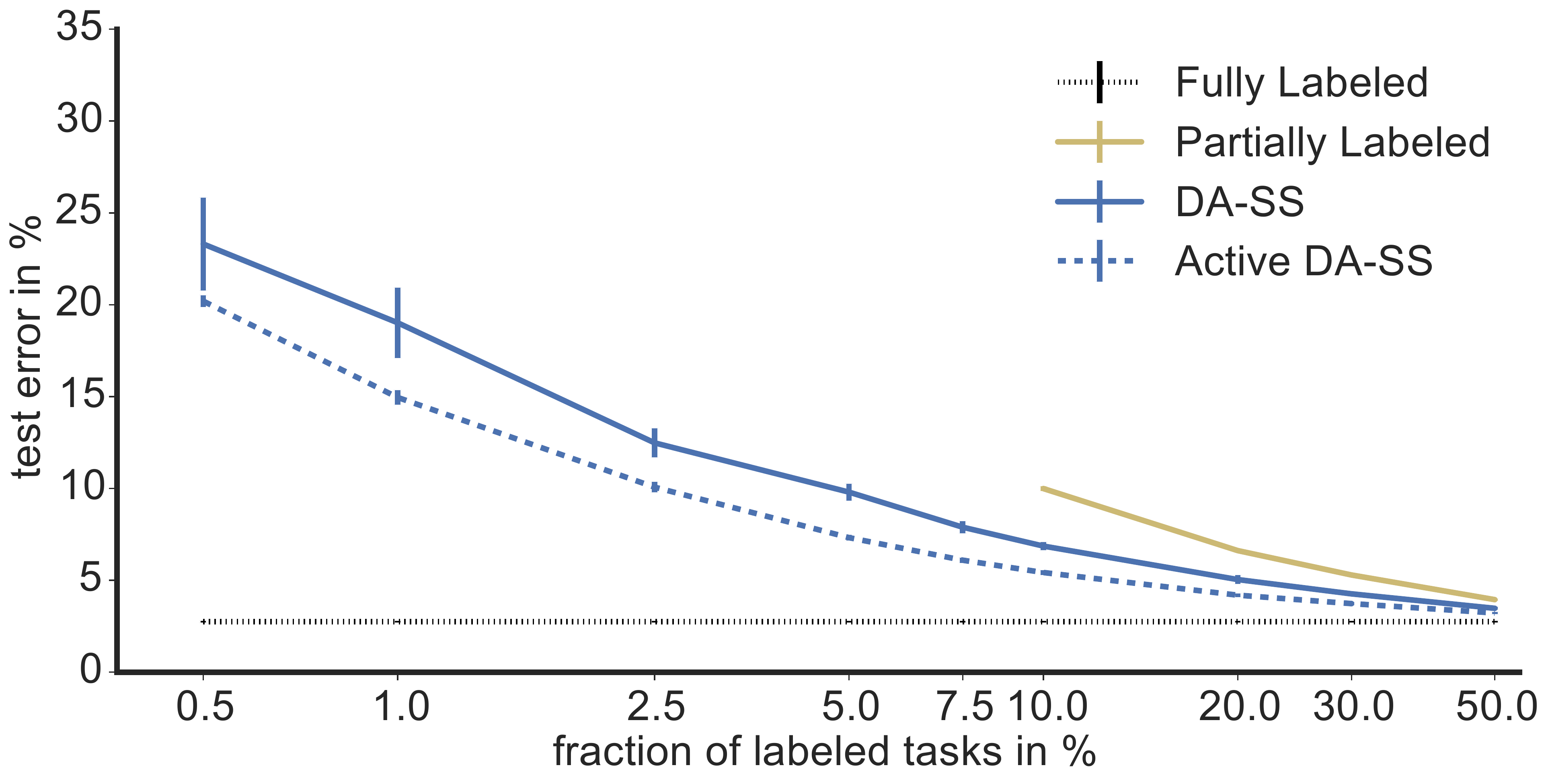}\quad}\label{fig:synthSS}
\subfigure[Product reviews, single-source transfer]{\quad\includegraphics[width=.4\textwidth]{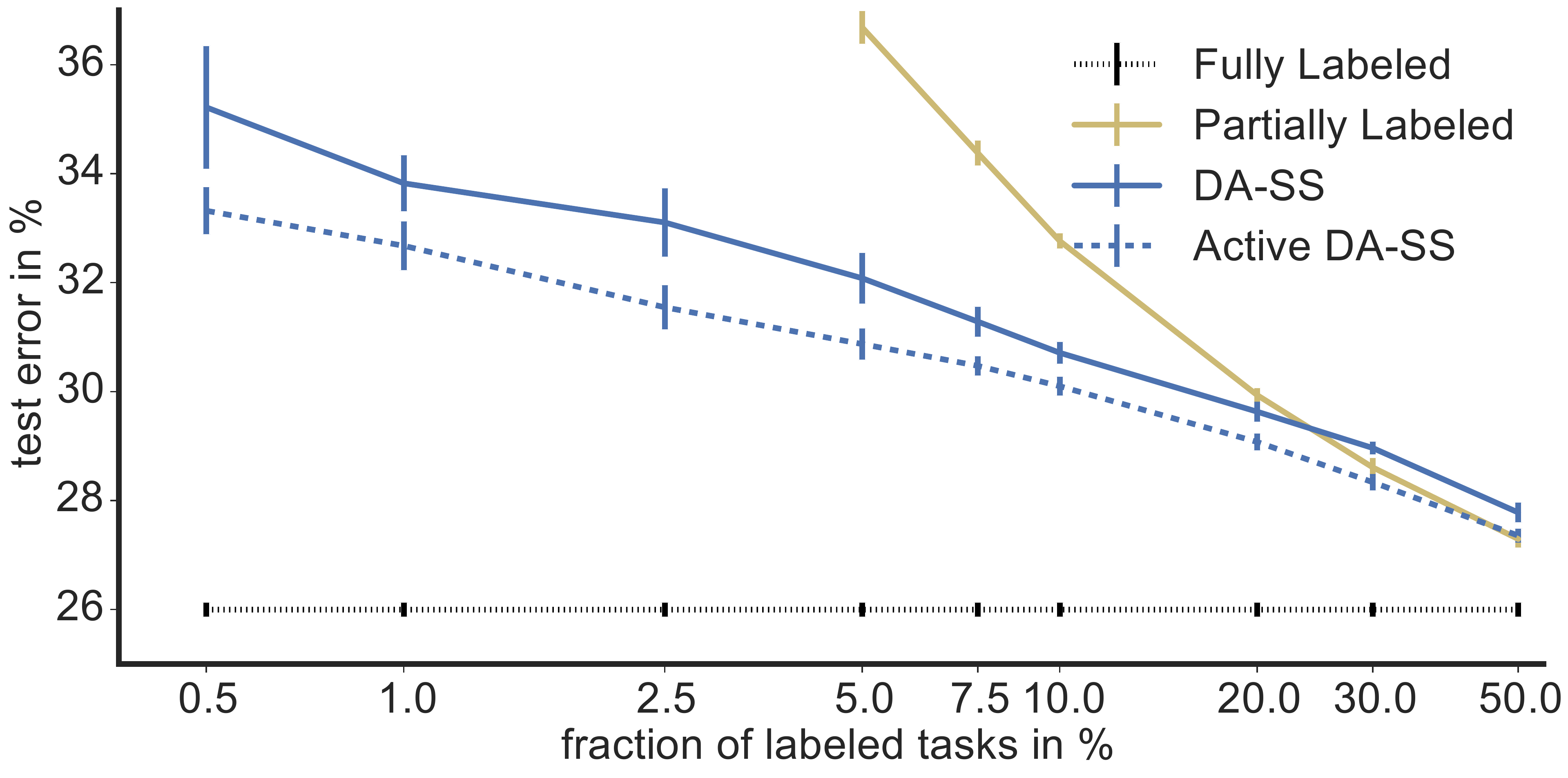}\quad}\label{fig:imagenetSS}
\caption{Experimental results on synthetic and real data: average test error and standard deviation over $10$ repeats.}
\end{figure*}

To illustrate that the proposed algorithm can also be practically useful, 
we performed experiments on synthetic and real data. 
In both cases we choose $\H$ to be the set of all linear predictors with a bias term on $\X=\mathbb{R}^d$.

\textbf{Synthetic data.}
We generate $T=1000$ binary classification tasks in $\mathbb{R}^2$.
For each task $t$ its marginal distribution $D_t$ is a unit-variance Gaussian with mean $\mu_t$ chosen uniformly at random from the set $[-5,5]\times[-5,5]$.
The label $+1$ is assigned to all points that have angle between $0$ and $\pi$ with $\mu_t$ (computed counter-clockwise), the other points are labeled $-1$.
We use $n=1000$ unlabeled and $m=100$ labeled examples per task.

\textbf{Real Data.} We curate a \emph{Multitask dataset of product reviews}\footnote{\scriptsize{\url{http://cvml.ist.ac.at/productreviews/}}} from the corpus of \emph{Amazon product data}\footnote{\scriptsize{\url{http://jmcauley.ucsd.edu/data/amazon/}}}~\cite{McAPanLes15,McATarShiHen15}.
We select the products for which there are at least $300$ positive reviews (with scores 4 or 5) and at least $300$ negative reviews (with scores 1 or 2).
Each of the resulting $957$ products we treat as a binary classification task of predicting whether a review is positive or negative.
For every review we extract features by first pre-processing (removing all non-alphabetical characters, transforming the rest into lower case and removing stop words) and then applying the sentence embedding procedure of~\cite{sentence} using $25$-dimensional GloVe word embedding~\cite{pennington2014glove}.
We use $n=500$ unlabeled samples per task and label a subset of $m=400$ examples for each of the selected tasks.
The remaining data is used for testing. 

\textbf{Methods.}  
We evaluate the proposed method in the case when the set of labeled tasks is predefined (referred to as \emph{DA}) by setting the set $I$ to be a random subset of tasks and minimizing~\eqref{eq:objective} only with respect to $\alpha$-s and in the active task selection scenario where~\eqref{eq:objective} is minimized with respect to both $I$ and $\alpha$-s (referred to as \emph{Active DA}).
We compare these methods to a multi-task method based on~\cite{ECCV12_Khosla}, also with random labeled tasks (the same ones as for \emph{DA}).
Specifically, we solve:
\begin{align}
\notag
\!\!\min_{w,v,b}C&\Big(\!\|w\|^2\!+\!\frac{1}{k}\!\sum_{t\in I}\!\|v_t\|^2\!\Big)\!+\!\frac{1\!-\!\gamma}{km}\!\!\!\!\!\sum_{t\in I, (x,y)\in\overline{S}_t}\!\!\!\!\!(w^T\!x\!+\!b\!-\!y)^2
\\
+\frac{\gamma}{km}&\sum_{t\in I}\sum_{(x,y)\in\overline{S}_t}((w^T+v_t^T)x+(b+b_t)-y)^2
\label{eq:objective-khosla}
\end{align}
for $\gamma\in\{0, 0.1,\dots, 1\}$ and use $(w,b)$ for making predictions on all unlabeled tasks and $(w+v_t,b+b_t)$ for each labeled task $t\in I$. 
For every number of labeled tasks we report the result for $\gamma$ that has the best test performance averaged over $10$ repeats (denoted by \emph{Multi-task}), as an upper performance bound on what could be achieved by model selection.

We also evaluate the discussed simplification of the proposed methods that consists of minimizing~\eqref{eq:objective2}.
We refer to these as \emph{DA-SS} (for random predefined labeled tasks) and as \emph{Active DA-SS} (in the active task selection scenario). The SS stands for \emph{single source}, as in this setting, each task is solved based on 
information from only one labeled tasks. 
 
To provide further context for the results we also report the results of learning independent ridge regressions with access to labels for all tasks (denoted by \emph{Fully Labeled}).
However, this baseline has access to many more annotated examples in total than all other methods.
In order to quantify this effect we also consider the setting when the learner has access to labels for all tasks, but fewer of them: namely, when the number of labeled tasks is $k$, the number of labels per task is $mk/T$, \ie the total amount of labeled examples is $mk$, the same as for all other methods.
In this case we evaluate two methods.
The first one learns ridge regressions for every task independently and thus can be seen as a reference point for the methods that do not involve information transfer between the labeled tasks, \ie \emph{DA-SS} and \emph{Active DA-SS}.
The second reference method is based on~\cite{Evgeniou} and consists of minimizing~\eqref{eq:objective-khosla} with $\gamma$ set to $1$ and processing all tasks as labeled.
This approach transfers information between all the tasks and therefore we refer to it when evaluating the methods that involve information transfer between the labeled tasks, \ie \emph{DA}, \emph{Active DA} and \emph{Multi-task}.

\textbf{Implementation.} We estimate the empirical discrepancies between pairs of tasks  by finding a hypothesis in $\H$ that minimizes the squared loss for the binary classification problem of separating the two sets of instances, as in~\cite{Ben-David:2010}. 
To minimize~\eqref{eq:objective} for a given set of labeled tasks we use gradient descent.
It is also used as a subroutine when minimizing~\eqref{eq:objective} with respect to both $I$ and $\alpha$-s, for which we employ the GraSP algorithm~\cite{journals/jmlr/BahmaniRB13}.
\emph{Active DA-SS} involves the minimization of the $k$-medoid risk~\eqref{eq:objective2}, which 
we perform using a local search as in~\cite{Park:2009:SFA:1464526.1465112}.
For both methods for the active task selection scenario we used the heuristic from $k$-means++ \cite{1283494} for initialization.
To obtain classifiers for the individual tasks in all scenarios we use least-squares ridge regression.
Regularization constants for all methods we selected from the set $\{0\}\cup\{10^{-17}, 10^{-16}\dots10^{8}\}$ by $5\times5$-fold cross validation.

%

\textbf{Results.} 
The results are shown in Figure~\ref{fig:experiments}. 
First, one can see that the proposed domain adaptation-inspired method \emph{DA} outperforms the multi-task method~\eqref{eq:objective-khosla}.
This could be due to higher flexibility of \emph{DA} compared to \emph{Multi-task} as the latter 
provides only one predictor for all unlabeled tasks.
Indeed, the difference is most apparent in the experiment with synthetic data, where by design there is no single predictor that could perform well on a large fraction of tasks.
Results on the product reviews indicate that \emph{DA}'s flexibility of learning a specific predictor for every task can be advantageous in more realistic scenarios as well.

Second, on both datasets both methods for active task selection, \ie \emph{Active DA} and \emph{Active DA-SS}, outperform the corresponding passive methods, \ie \emph{DA} and \emph{DA-SS}, systematically across various fractions of the labeled tasks.
In particular, both active task selection methods require substantially fewer tasks labeled to achieve the same accuracy as their analogs with randomly chosen tasks.
This confirms the intuition that selecting which tasks to label in a data-dependent way is beneficial and  demonstrates that Theorem~\ref{thm:main} is capable of capturing this effect.

Another interesting observation that can be made from the results in Figure~\ref{fig:experiments} is that both active and passive domain adaptation-inspired methods clearly outperform the corresponding partially labeled baselines, especially for small fractions of labeled tasks.
This indicates that having more labels for fewer tasks rather than only few labels for all tasks could be beneficial not only in terms of annotation costs, but also in terms of prediction accuracy. 
%

As the number of labeled tasks gets larger, \eg half of all tasks, the performance of the active task 
selection learner becomes almost identical to the performance of the \emph{Fully Labeled} method, 
even improving over it in the case of multi-source transfer on synthetic data. 
This confirms the intuition that in the case of many related tasks even a fraction of the tasks 
can contain enough information for solving all tasks. 

\section{Conclusion}

In this work we introduced and studied a variant of multi-task learning in which annotated data is available only for some of the tasks.
This setting combines aspects of traditional multi-task learning, namely the transfer of information between labeled tasks, with aspects typical for domain adaptation problems, namely transferring information from labeled tasks to solve tasks for which only unlabeled data is available.
The success of the learner in this setting depends on the effectiveness of information transfer and informativeness of the set of labeled tasks.
We analyzed two scenarios: a passive one, in which the set of labeled tasks is predefined, and the \emph{active task selection} scenario, in which the learner decides for which tasks to query labels.

Our main technical contribution is a generalization bound that quantifies the informativeness of the set of labeled tasks and the effectiveness of information transfer. 
We demonstrated how the bound can be used to make the choice of labeled tasks (in the active scenario) and to transfer information between the tasks in a principled way.
We also showed how the terms in the bound have intuitive interpretations and provide guidance under which assumption of tasks relatedness the induced algorithm is expected to work well. 
Our empirical evaluation demonstrated that the proposed methods work also well in practice.

For future work we plan to further exploit the idea of active learning in the multi-task setting. 
In particular, we are interested in identifying whether by allowing the learner to make its decision on which tasks to label in an iterative way, rather than forcing it to choose all the tasks at the same time, one could obtain better learning guarantees as well as more effective learning methods. 

\section*{Acknowledgments} 
We thank Alexander Zimin and Marius Kloft for useful discussions.
This work was in parts funded by the European Research Council under the European Union's 
Seventh Framework Programme (FP7/2007-2013)/ERC grant agreement no 308036.

\bibliographystyle{plain}

\appendix

\section{Preliminaries}

In this section we list a few results from the literature that will be utilized in the proof of Theorem~\ref{thm:main}.

\begin{proposition}[Lemma 1 in~\cite{Ben-David:2010}]\label{thm:disc-estimation}
Let $d$ be the VC dimension of the hypothesis set $\H$ and $S_1,S_2$ be two \iid samples of size $n$ from $D_1$ and $D_2$ respectively. Then for any $\delta>0$ with probability at least $1-\delta$:
\begin{align*}
\disc(D_1,D_2)&\!\leq\!\disc(S_1,S_2)\!+\!2\sqrt{\frac{2d\log(2n)+\log(2/\delta)}{n}}.
\end{align*}
\end{proposition}

\begin{lemma}[Theorem 1 in~\cite{Maurer:2006}]
Let $X_1,\dots,X_n$ be independent random variables taking values in the set $\X$ and $f$ be a function $f:\X^n\rightarrow\mathbb{R}$. For any $x=(x_1,\dots,x_n)\in\X^n$ and $y\in\X$ define:
\begin{align*}
x_{y,k}&=(x_1,\dots,x_{k-1},y,x_{k+1},\dots,x_n)\\
(\inf_k f)(x)&=\inf_{y\in\X}f(x_{y,k})\\
\Delta_{+,f}&=\sum_{i=1}^n(f-\inf_k f)^2.
\end{align*}
Then for $t>0$:
\begin{equation}
\Pr\{f-\E f\geq t\}\leq\exp\left(\frac{-t^2}{2\|\Delta_{+}\|_{\infty}}\right).
\end{equation}
\label{lem:maurer}
\end{lemma}

\begin{lemma}[Corollary 6.10 in~\cite{mcdiarmid89}] Let $W_0^n$ be a martingale with respect to a sequence of random 
variables $(B_1,\dots,B_n)$. Let $b_1^n=(b_1,\dots,b_n)$ be a vector of possible values of the random variables $B_1,\dots,B_n$. 
Let
\begin{equation}
r_i(b_1^{i-1})=\sup_{b_i}\{W_i: B_1^{i-1}=b_1^{i-1}, B_i=b_i\}-\inf_{b_i}\{W_i: B_1^{i-1}=b_1^{i-1}, B_i=b_i\}.
\end{equation}
Let $r^2(b_1^n)=\sum_{i=1}^n(r_i(b_1^{i-1}))^2$ and $\widehat{R}^2=\sup_{b_1^n}r^2(b_1^n)$. Then
\begin{equation}
\Pr_{B_1^n}\{W_n-W_0>\epsilon\}<\exp\left(-\frac{2\epsilon^2}{\widehat{R}^2}\right).
\end{equation}
\label{lem:mcdiarmid}
\end{lemma}

\begin{lemma}[Originally~\cite{Hoeffding:1963}; in this form Theorem 18 in~\cite{TolBlaKlo14}]
Let $\{U_1,\dots,U_m\}$ and $\{W_1,\dots,W_m\}$ be sampled uniformly from a finite set of $d$-dimensional vectors $\{v_1,\dots,v_N\}\subset\mathbb{R}^d$ with and without replacement respectively. Then for any continuous and convex function $F:\mathbb{R}^d\rightarrow\mathbb{R}$ the following holds:
\begin{equation}
\E\left[F\left(\sum_{i=1}^mW_i\right)\right]\leq\E\left[F\left(\sum_{i=1}^mU_i\right)\right]
\end{equation}
\label{lem:hoef_convexity}
\end{lemma}

\begin{lemma}[Part of Lemma 19 in~\cite{TolBlaKlo14}]
Let $x=(x_1,\dots,x_l)\in\mathbb{R}^l$. Then the following function is convex:
\begin{equation}
F(x)=\sup_{i=1\dots l}x_i.
\end{equation}
\label{lem:convexity}
\end{lemma}

\section{Proof of Theorem ~\ref{thm:main}}
\label{app:proof}
We start with bounding the multi-task error by the errors on the source tasks, and transition to empirical quantities while keeping the effect of random sampling controlled.

Fix a subset of labeled tasks $I=\{i_1,\dots,i_k\}$, a task $\langle D_t,f_t\rangle$ and a weight vector $\alpha\in\Lambda^I$.
Let $h^*_i\in\arg\min_{h\in \H}(\er_t(h)+\er_i(h))$.\footnote{If the minimum is not attained, the same inequality follows by an argument of arbitrary close approximation.} 
Writing $\ell(h,h')$ as shorthand for $\ell(h(x),h'(x))$, we have
\begin{align}
|\er_\alpha(h)&-\er_t(h)|=\Big|\sum_{i\in I}\alpha_i\er_i(h)-\er_{t}(h)\Big|\leq\sum_{i\in I}\alpha_i\big|\er_{i}(h)-\er_{t}(h)\big|
\\
&\leq\sum_{i\in I}\!\alpha_i\!\left(\big|\er_{i}(h)-\!\!\!\E_{x\sim D_i}\!\!\ell(h,h^\ast_i)\big|
+\big|\E_{x\sim D_i}\!\!\ell(h,h^\ast_i)-\!\!\!\E_{x\sim D_t}\!\!\ell(h,h^\ast_i)\big| +\big|\er_{t}(h)-\!\!\!\E_{x\sim D_t}\!\!\ell(h,h^\ast_i)\big|\right)=(*)
\end{align}
We can bound each summand: 
\begin{align*}
|\er_{i}(h)-\underset{x\sim D_i}{\E}\!\!\!\ell(h,h^\ast_i)|&\leq \er_i(h^\ast) \\
|\E_{x\sim D_i}\!\ell(h,h^\ast_i)-\E_{x\sim D_t}\!\ell(h,h^\ast_i)\big|&\leq \disc(D_i,D_t)\\
|\er_t(h)-\!\!\E\limits_{x\sim D_t}\!\!\ell(h,h^\ast_i)\big| &\leq \er_t(h^*_i)
\end{align*}
where the first and the last inequalities hold by the  triangular inequality for $\ell$ and the second one follows from the definition of discrepancy.
Therefore,
\begin{align}
(*)\leq
&\sum_{i\in I}\alpha_i(\er_{i}(h^*_i)+\disc(D_i,D_t)+\er_{t}(h^*_i))=\sum_{i\in I}\alpha_i(\lambda_{it}+\disc(D_i,D_t)).
\end{align}
Consequently, assuming that every task $t$ has its own weights $\alpha^t$ we obtain that:
\begin{align}
\frac{1}{T}\sum_{t=1}^T\er_t(h)\leq\frac{1}{T}\sum_{t=1}^T\er_{\alpha^t}(h_t)+\frac{1}{T}\sum_{t=1}^T\sum_{i\in I}\alpha^t_{i}\disc(D_t,D_{i})+\frac{1}{T}\sum_{t=1}^T\sum_{i\in I}\alpha^t_{i}\lambda_{ti}.
\label{eq:traingular-ineq-ms}
\end{align}
We continue with bounding every expectation on the right hand side of~\eqref{eq:traingular-ineq-ms} by its empirical counterpart.

\subsection{ Bound $\frac{1}{T}\sum_{t=1}^T\sum_{i\in I}\alpha^t_{i}\disc(D_t,D_{i})$}

We apply Proposition~\ref{thm:disc-estimation} to every summand and combine the results using a union bound argument.
We obtain that with probability at least $1-\delta/2$ uniformly for all choices of $I$ and $\alpha^1,\dots,\alpha^T\in\Lambda^I$:
\begin{align}
\frac{1}{T}\sum_{t=1}^T\sum_{i\in I}\alpha^t_{i}\disc&(D_t,D_{i})\leq\frac{1}{T}\sum_{t=1}^T\sum_{i\in I}\alpha^t_i\disc(S_t,S_{i})+2\sqrt{\frac{2d\log(2n)+\log(4T^2/\delta)}{n}}.
\label{eq:bound-disc-ms}
\end{align} 

\subsection{ Bound $\frac{1}{T}\sum_{t=1}^T\er_{\alpha^t}(h_t)$}

Now we upper-bound the error term in two steps. 

\subsubsection{ Relate $\frac{1}{T}\sum_{t=1}^T\er_{\alpha^t}(h_t)$ to $\frac{1}{T}\sum_{t=1}^T\tilde{\er}_{\alpha^t}(h_t)$}

We start with relating the multi-task error to the hypothetical empirical error, if the learner would receive labels for all examples in the selected labeled tasks:
\begin{align}
\tilde{\er}_{\alpha}(h)&=\sum_{i\in I}\alpha_i\eer_{S^u_i}(h)\\
\intertext{for}
\quad \eer_{S^u_i}(h)&=\frac{1}{n}\sum_{j=1}^n\ell(h(x^i_j),f_i(x^i_j)).
\end{align}
Clearly, if $m=n$ this part is not necessary and we can avoid the resulting complexity terms.

Because the choice of the tasks to label, $I$, their weights, $\balpha=(\alpha^1,\dots,\alpha^T)$, and the predictors, $\bh=(h_1,\dots,h_T)$, all depend on the unlabeled data, we aim for a bound that is holds simultaneous for all choices of these quantities, under the condition that $I$ and $\balpha$ depend only on the unlabeled samples, while $\bh$ can be chosen based also on the labeled subsets.

Our main tool is a refined version of McDiarmid's inequality, due to Maurer~\cite{Maurer:2006} (Lemma~\ref{lem:maurer}), which allows us to make use of the internal structure of the weights, $\balpha$, while deriving a large deviation bound.

For any $\bS=(S^u_1,\dots,S^u_T)$ 
define: 
\begin{align}
\Psi(\bS) & =\sup_{I=\{i_1,\dots,i_k\}}\sup_{\alpha^1,\dots,\alpha^n\in\Lambda^I}\sup_{h_1,\dots,h_T}
\frac{1}{T}\sum_{t=1}^T\sum_{i=1}^T\alpha^t_{i}(\er_{i}(h_t)-\eer_{S^u_i}(h_t))
 =\ \sup_I\sup_{\balpha}\sup_{\bh}\ g(\balpha,\bh,\bS) \label{eq:Psi-def}
\intertext{ for}
g(\balpha,\bh,\bS)& =\sum_{i=1}^T\sum_{j=1}^n\left(\frac{1}{Tn}\sum_{t=1}^T\alpha^t_i(\er_i(h_t)-\ell(h_t(x^i_j),f_t(x^i_j)))\right).
\end{align}
For notational simplicity we will sometimes think of every $S^u_t$ as a set of \emph{pairs} $(x^t_i,y^t_i)$, where $y^t_i=f_t(x^t_i)$.
To apply Lemma~\ref{lem:maurer} we establish a bound on $\Delta_{+,\Psi}(\bS)=\sum_i\sum_j(\Psi(\bS)-\Psi_{ij}(\bS))^2$,
with
\begin{equation}
\Psi_{ij}(\bS)=\inf_{(x,y)}\sup_{\balpha}\sup_{\bh} \ g(\balpha,\bh, \bS\setminus \{(x^i_j, y^i_j)\}\cup\{(x,y)\},
\end{equation}
\ie the possible smallest value for $\Psi$ when changing only the data point $(x^i_j, y^i_j)$.
Let $\balpha^*,\bh^*$ be the point where the $\sup$ in the~\eqref{eq:Psi-def} is attained\footnote{If the 
supremum is not attained the subsequent inequality still follows from an argument of arbitrarily close approximation.},
\ie $\Psi(\bS)=g(\balpha^*,\bh^*,\bS)$. Then:
\begin{equation}
\Psi_{ij}(\bS)\geq\inf_{(x,y)}g(\balpha^*,\bh^*,\bS\setminus \{(x^i_j, y^i_j)\}\cup\{(x,y)\}\ )
\end{equation}
and therefore
\begin{align}
\Psi(\bS)-\Psi_{ij}(\bS)
&\leq g(\balpha^*,\bh^*,\bS) - \inf_{(x,y)}g(\balpha^*,\bh^*,\bS\setminus \{(x^i_j, y^i_j)\}\cup\{(x,y)\})
\\
&\leq \sup_{(x,y)}\frac{1}{Tn}\sum_{t=1}^T\alpha^{*t}_i(-\ell(h^*_t(x^i_j),y^i_j)+\ell(h^*_t(x),y))
\ \leq\ \frac{1}{Tn}\sum_{t=1}^T\alpha^{*t}_i,
\end{align}
where for the last inequality we use that $\ell$ is bounded in $[0,1]$.
Because also $\Psi(\bS)-\Psi_{ij}(\bS) \geq 0$, we obtain
\begin{align}
\Delta_{+,\Psi}(\bS) &= \sum_{i=1}^T\sum_{j=1}^n(\Psi(\bS)-\Psi_{ij}(\bS))^2\leq \sum_{i=1}^T\sum_{j=1}^n\frac{1}{T^2n^2}\left(\sum_{t=1}^T\alpha^{*t}_i\right)^2
\leq \frac{1}{T^2n}\left(\sum_{i=1}^T\sum_{t=1}^T\alpha^{*t}_i\right)^2=\frac{1}{n},
\end{align}
(remember that $\sum_i \alpha_i=1$ for any $\alpha\in\Lambda^I$). Therefore, according to Lemma~\ref{lem:maurer} with probability at least $1-\delta/4$:
\begin{align}
\Psi(\bS)\leq\E\,\Psi(\bS)+\sqrt{\frac{2}{n}\log\frac{4}{\delta}}. \label{eq:step31-ms-first-half}
\end{align}
To bound $\E_S\Psi(\bS)$ we use symmetrization and Rademacher variables, $\sigma_{ij}$:
\begin{align}
\E_{\bS}\Psi(\bS) &=\E_{\bS}\ \sup_I\sup_{\alpha^1,\dots,\alpha^T\in\Lambda^I}\sup_{h_1,\dots,h_T}\ 
\sum_{i=1}^T\sum_{j=1}^n\left(\frac{1}{Tn}\sum_{t=1}^T\alpha^t_i(\er_i(h_t)-\ell(h_t(x^i_j),y^i_j))\right)
\\
&\leq \label{line3}
2\E_{\bS}\E_\sigma\sup_I\sup_{\alpha^1,\dots,\alpha^T\in\Lambda^I}\sup_{h_1,\dots,h_T}\sum_{i=1}^T\sum_{j=1}^n\left(\frac{\sigma_{ij}}{Tn}\sum_{t=1}^T\alpha^t_i\ell(h_t(x^i_j),y^i_j)\right)
\\ &\leq
\label{line4}
2\E_{\bS}\E_\sigma\ \frac{1}{T}\sum_{t=1}^T\sup_{\alpha^t\in\Lambda,h_t}\sum_{i=1}^T\sum_{j=1}^n\frac{\sigma_{ij}\alpha^t_i}{n}\sum_{t=1}^T\ell(h_t(x^i_j),y^i_j)
\\ &\leq
2\E_{\bS}\E_\sigma\ \ \sup_{\alpha,h}\sum_{i=1}^T\sum_{j=1}^n\frac{\sigma_{ij}\alpha_i}{n}\ell(h(x^i_j),y^i),
\label{eq:linearinalpha}
\end{align}
where line~\eqref{line4} is obtained from line~\eqref{line3} by dropping the assumption of a common sparsity pattern between the $\alpha$-s.
Note that the function inside the last $\sup$ is linear in $\alpha\in\Lambda$, therefore $\sup_\alpha$ can be reduced to the $\sup$ 
over the corners of the simplex, $\{(1,0,\dots,0),\,\dots\,,(0,\dots,0,1)\}$.
At the same time, by Sauer's lemma, the number of different choices of $h$ on $\bS$ is bounded by $\left(\frac{eTn}{d}\right)^{d}$.
Therefore, the total number of different choices in~\eqref{eq:linearinalpha} is bounded by
$T\left(\frac{enT}{d}\right)^{d}$.
%
Furthermore, for any choice of $\alpha$ and $h$, the norm of the $Tn$-vector formed by the summands of~\eqref{eq:linearinalpha}
is bounded by $1/\sqrt{n}$, because
\begin{align}
\sum_{i=1}^T\sum_{j=1}^n \left(\frac{\sigma_{ij}\alpha_i}{n}\ell(h(x^i_j),y^i)\right)^2 
&= \frac{1}{n^2}\sum_{i=1}^T\sum_{j=1}^n \left(\alpha_i\ell(h(x^i_j),y^i)\right)^2 
\leq \frac{1}{n^2}\sum_{j=1}^n \left(\sum_{i=1}^T\alpha_i\right)^2 = \frac{1}{n}.
\end{align}
Therefore, by Massart's lemma:
\begin{align}
\E_\sigma\ \ \sup_{\alpha,h}\sum_{i=1}^T\sum_{j=1}^n\frac{\sigma_{il}\alpha_i}{n}\ell(h(x^i_l),y^i_l)\leq\frac{\sqrt{2(\log T +d\log(enT/d))}}{\sqrt{n}}. \label{eq:step31-ms-second-half}
\end{align}

Combining~\eqref{eq:step31-ms-first-half} and \eqref{eq:step31-ms-second-half} we obtain that 
with probability at least $1-\delta/4$ simultaneously for all choices of tasks 
to be labeled, $I$, weights $\balpha$ and hypotheses $\bh$:
\begin{align}
\frac{1}{T}\sum_{t=1}^T\er_{\alpha^t}(h_t)\leq\frac{1}{T}\sum_{t=1}^T\tilde{\er}_{\alpha^t}(h_t)+\sqrt{\frac{8(\log T+d\log(enT/d))}{n}}+\sqrt{\frac{2}{n}\log\frac{4}{\delta}}.
\label{eq:step31-ms}
\end{align}
\subsubsection{ Relate $\frac{1}{T}\sum_{t=1}^T\eer_{\alpha^t}(h_t)$ to $\frac{1}{T}\sum_{t=1}^T\tilde{\er}_{\alpha^t}(h_t)$}

Fix the unlabeled samples $S^u_1,\dots, S^u_T$. This uniquely 
determines the chosen tasks $I$ and the weights $\alpha^1,\dots,\alpha^T\in\Lambda^I$, so the only remaining source of randomness is the uncertainty which subsets of 
the selected tasks are labeled. 
%
%

For notational simplicity we pretend that exactly the first $k$ tasks were selected,
\ie $I=\{1,\dots,k\}$. The general case can be obtained by changing the indices in 
the proof from $1,\dots,k$ to $i_1,\dots,i_k$. 

To deal with the dependencies between the labeled data points we first note 
that any random labeled subset $S^l_i=(\bar s^i_1,\dots,\bar s^i_m)$ can be 
described as the first $m$ elements of a random permutation $Z_i=(z^i_1,\dots,z^i_n)$ 
over $n$ elements that correspond to the unlabeled sample $S^u_i$, \ie
$\bar s^i_j=(\bar x^i_j,\bar y^i_j)=(x^i_{z^i_j},y^i_{z^i_j})$.
With this notation and writing $\bZ=(Z_1,\dots,Z_k)$ and $\ell(h, z^i_j)=\ell(h(\bar x^i_j),\bar y^i_j)$ we define the following function
\begin{align}
\Phi(\bZ)&=\sup_{h_1,\dots,h_T}\frac{1}{T}\sum_{t=1}^T\tilde{\er}_{\alpha^t}(h_t)-\eer_{\alpha^t}(h_t)
=\sup_{h_1,\dots,h_T}\sum_{i=1}^k\frac{1}{T}\sum_{t=1}^T\alpha^t_i\Big(
\frac{1}{n}\sum_{j=1}^n\ell(h_t,z^i_j) 
-\frac{1}{m}\sum_{j=1}^m\ell(h_t,z^i_j)\Big).
\end{align}
%
Our main tool is McDiarmid's inequality (Lemma~\ref{lem:mcdiarmid}) for martingales. 

\paragraph{Construct a martingale sequence}\mbox{}\\

For this, we interpret $\bZ=(z^1_1,z^1_2,\dots,z^k_n)$ as a sequence of $kn$ dependent 
variables, $z_{11},\dots,z_{kn}$. 
For the sake of notational consistency we will keep using double indices, with 
the convention that the sample index, $j=1,\dots,n$, runs faster than the task 
index, $i=1,\dots,k$. Segments of a sequence will be denoted by upper and lower double 
indices, $z^{\bar\imath\bar\jmath}_{ij} = (z_{ij},z_{i(j+1)},\dots,z_{\bar\imath\bar\jmath})$ for $ij\leq \bar\imath\bar\jmath$ and 
$z^{\bar\imath\bar\jmath}_{ij}=\emptyset$ otherwise.
We now create a martingale sequence using Doob's construction~\cite{doob1940regularity}:
\begin{equation}\label{eq:Wij}
W_{ij}=\E_{\bZ}\{\Phi(\bZ)|\ z_{11}^{ij}\ \}.
\end{equation}
where here and in the following when taking expectations over $\bZ$ 
it is silently assumed that the expectation is taken only with respect to variables 
that are not conditioned on. 
Note that because of this convention, the expectations in \eqref{eq:Wij} is only 
with respect to $z_{i(j+1)},\dots,z_{kn}$, so each $W_{ij}$ is a random variable 
of $z_{11},\dots,z_{ij}$. In particular, $W_{00} = \E_{\bZ}\Phi(\bZ)$ and $W_{kn}=\Phi(\bZ)$,
and the in between sequence is a martingale with respect to $z_{11},\dots,z_{kn}$: 
\begin{align}
\E_{\bZ}\{\ W_{ij} | z_{11}^{i(j-1)}\ \} = 
\E_{\bZ}\big\{\ \E_{\bZ}\{\Phi(\bZ)|\ z_{11}^{ij}\} \big|\ z_{11}^{i(j-1)}\ \big\} = 
\E_{\bZ}\{\Phi(\bZ)| z_{11}^{i(j-1)}\ \} = W_{i(j-1)}.
\end{align}
\newline
\paragraph{Upper-bound $\widehat{R}^2$}\mbox{}\\

In order to apply Lemma~\ref{lem:mcdiarmid} we need an upper bound on the coefficient $\widehat{R}^2$ defined there. 
%

Let $i\in\{1,\dots,k\}$ and $j\in\{1,\dots,n\}$ be fixed and let $\bpi=(\pi_1,\dots,\pi_k)$ 
be specific permutations of $n$ elements for which we use the same 
index conventions as for $\bZ$. 
By $\sigma$ and $\tau$ will denote elements in $\pi^{in}_{i(j+1)}$, 
\ie $\sigma$ and $\tau$ do not occur in any of the first $j$ positions 
of the permutation $\pi_i$. 
Then
\begin{align}
\notag
r_{ij}(\pi_{11}^{i(j-1)})
&=
 \sup_{\sigma\in\pi^{in}_{i(j+1)}}\{\ W_{ij}: z_{11}^{i(j-1)} = \pi_{11}^{i(j-1)},\ z_{ij}=\sigma\}
-\inf_{\sigma\in\pi^{in}_{i(j+1)}}\!\!\! \{\ W_{ij}: z_{11}^{i(j-1)} = \pi_{11}^{i(j-1)},\ z_{ij}=\sigma\}
\\
&=\sup_{\sigma\in\pi^{in}_{i(j+1)}}\sup_{\tau\in\pi^{in}_{i(j+1)}}\!\!\!\Big[
\E_{z_{i(j+1)}^{kn}} \{\Phi(\pi_{11}^{i(j-1)},\sigma, z_{i(j+1)}^{kn})\}
-
\!\!\!\E_{z_{i(j+1)}^{kn}} \{\Phi(\pi_{11}^{i(j-1)},\tau, z_{i(j+1)}^{kn})\}\Big].
\label{line2}
\end{align}
To analyze~\eqref{line2} further, recall that:
\begin{align*}
&\E_{z_{i(j+1)}^{kn}} \{\Phi(\pi_{11}^{i(j-1)},\sigma, z_{i(j+1)}^{kn})\}
=
\sum_{\pi_{i(j+1)}^{kn}}\Phi(\pi_{11}^{i(j-1)},\sigma, \pi_{i(j+1)}^{kn})
\times
\Pr(\ z_{i(j+1)}^{kn}=\pi_{i(j+1)}^{kn}\ |
 z_{11}^{i(j-1)}=\pi_{11}^{i(j-1)} \wedge z_{ij}=\sigma\ ),
 \end{align*}
where here and in the following we use the convention that sums 
over parts of $\bpi$ run only over values that lead to valid permutations. 
Because the permutations of different task are independent, this is equal to
\begin{align}
\label{eq:line5}
=
\sum_{\pi_{i(j+1)}^{kn}}\Phi(\pi_{11}^{i(j-1)},\sigma, \pi_{i(j+1)}^{kn}
\Pr(\ z_{i(j+1)}^{in}=\pi_{i(j+1)}^{in}\ |
 z_{i1}^{i(j-1)}=\pi_{i1}^{i(j-1)} \wedge z_{ij}=\sigma\ )
\Pr(z_{(i+1)1}^{kn}=\pi_{(i+1)1}^{kn})
\end{align}
%
%
%
We make the following observation: for any fixed $\pi_{i1}^{ij}$ and any $\tau\not\in \pi_{i1}^{ij}$,
we can rephrase a summation over $\pi_{i(j+1)}^{in}$ into a sum over all positions where $\tau$ 
can occur, and a sum over all configuration for the entries that are not $\tau$:
\begin{align} 
\sum_{\pi_{i(j+1)}^{in}} F(\pi_{i(j+1)}^{in}) &= 
\sum_{l=j+1}^n \sum_{\pi_{i(j+1)}^{i(l-1)}}\sum_{\pi_{i(l+1)}^{in}} F(\pi_{i(j+1)}^{i(l-1)},\tau,\pi_{i(l+1)}^{in})
\end{align}
for any function $F$. Applying this to the summation in \eqref{eq:line5}, we obtain 
\begin{align*}
&\sum_{\pi_{i(j+1)}^{kn}}\Phi(\pi_{11}^{i(j-1)},\sigma, \pi_{i(j+1)}^{kn})
\Pr(\ z_{i(j+1)}^{in}=\pi_{i(j+1)}^{in}\ |
 z_{i1}^{i(j-1)}=\pi_{i1}^{i(j-1)} \wedge z_{ij}=\sigma\ )\\
 &\quad\times\Pr(z_{(i+1)1}^{kn}=\pi_{(i+1)1}^{kn})=\sum_{l=j+1}^n \sum_{\pi_{i(j+1)}^{i(l-1)}}\sum_{\pi_{i(l+1)}^{kn}}
\Phi(\pi_{11}^{i(j-1)},\sigma, \pi_{i(j+1)}^{i(l-1)}, \tau, \pi_{i(l+1)}^{kn})
\end{align*}
 \begin{align*}
&\quad\times
\Pr(\,z_{i(j+1)}^{i(l-1)}=\pi_{i(j+1)}^{i(l-1)}\wedge z_{i(l+1)}^{kn}=\pi_{i(l+1)}^{kn} | z_{11}^{i(j-1)}=\pi_{11}^{i(j-1)}\wedge z_{ij}=\sigma \wedge z_{il}=\tau )\\
&\quad\times\Pr(z_{(i+1)1}^{kn}=\pi_{(i+1)1}^{kn})=\E_{l\sim U_{j+1}^n} \E_{\bZ}\ \Phi(\bZ|z_{11}^{i(j-1)}=\pi_{11}^{i(j-1)}\wedge z_{ij}=\sigma\wedge z_{il}=\tau),
\end{align*}
where $U_{j+1}^n$ denotes the uniform distribution over the set $\{j+1,\dots,n\}$. 
%
The analogue derivation can be applied to the quantity in line~\eqref{line2} with $\sigma$ and $\tau$ exchanged.

For any $\bZ$ denote by $\bZ^{ij\leftrightarrow il}$ the permutation obtained by switching $z_{ij}$ and $z_{il}$. 
Then, due to the linearity of the expectation:
\begin{align}
r_{ij}(\pi_{11}^{i(j-1)})
&=\sup_{\sigma,\tau}\ 
\{\E_{l\sim U_{j+1}^n}\E_{\bZ}\{\Phi(\bZ)-\Phi(\bZ^{ij\leftrightarrow il})|z^{i(j-1)}_{11}=\pi^{i(j-1)}_{11}, z_{ij}=\sigma, z_{il}=\tau).\label{eq:rij_difference}
\end{align}
From the definition of $\Phi$ we see that $\Phi(\bZ)-\Phi(\bZ^{ij\leftrightarrow il})=0$ 
when $j,l\in\{1,\dots,m\}$  or $j,l\in\{m+1\,\dots,n\}$. 
%
%
Since $l>j$ in \eqref{eq:rij_difference} this implies $r_{ij}(\pi_{11}^{i(j-1)})=0$ for $j\in\{m+1,\dots,n\}$.
The only remaining cases are $j\in\{1,\dots,m\}$ and $l\in\{m+1,\dots,n\}$, for which we obtain
\begin{align*}
\Phi(\bZ)-\Phi(\bZ^{ij\leftrightarrow il})
&\leq\sup_{h_1,\dots,h_T}\frac{1}{T}\sum_{t=1}^T\alpha_i^t\frac{1}{m}(-\ell(h_t,z^i_j)+\ell(h_t,z^i_l))\leq\frac{1}{Tm}\sum_{t=1}^T\alpha^t_i.
\end{align*}
where for the first inequality we used that $\sup F-\sup G \leq \sup(F-G)$ for any $F,G$, and for the second inequality we 
used that $\ell$ is bounded by $[0,1]$. 
Consequently, $r_{ij}(\pi_{11}^{i(j-1)})\leq\frac{n-m}{n-j}\frac{1}{Tm}\sum_{t=1}^T\alpha^t_i$ in this case.
Therefore\footnote{We generously bound $\frac{n-m}{n-j}\leq 1$ in this step. By keeping the corresponding factor 
in the analysis one obtains that the constant $B$ in the theorem can be improved at least by a factor of $\frac{(n-m)^2}{(n-0.5)(n-m-0.5)}$.}
\begin{align}
\widehat{R}^2
= \sum_{i=1}^k\sum_{j=1}^n \big(r_{ij}(\pi_{11}^{i(j-1)})\big)^2 
&\leq \frac{1}{T^2m^2} \sum_{j=1}^m\Big(\frac{n-m}{n-j}\Big)^2 \sum_{i=1}^k \left(\sum_{t=1}^T\alpha^t_i\right)^2\leq\frac{1}{T^2 m}\sum_{i=1}^k\left(\sum_{t=1}^T\alpha^t_i\right)^2.
\label{eq:step322}
\end{align}
%
%
\newline
\paragraph{ Upper-bound $\E_{\bZ}\Phi(\bZ)$}\mbox{}\\

The main tool here is Lemma~\ref{lem:hoef_convexity}.
First we rewrite $\Phi(\bZ)$ in the following way:
\begin{align*}
\Phi(\bZ)&=\frac{1}{T}\sum_{t=1}^T\sup_h\sum_{i=1}^k\alpha^t_i(\eer_{S^u_i}(h)-\eer_{S^l_i}(h))=\frac{1}{Tm}\sum_{t=1}^T\Phi_t(\bZ)\\
\Phi_t(\bZ)&=\sup_h\sum_{i=1}^k m\,\alpha^t_i(\eer_{S^u_i}(h)-\eer_{S^l_i}(h)).
\end{align*}
Note that even though $\H$ can be infinitely large, we can identify
a finite subset that represents all possible predictions of hypothesis 
in $\H$ on $S^u_1\cup\cdots\cup S^u_k$.
We denote their number by $L\leq 2^{kn}$ and the corresponding hypotheses by $h^1,\dots,h^L$.

Let $t\in\{1,\dots,T\}$ be fixed.
For every $i\in\{1,\dots,k\}$ define a set 
of $n$ $L$-dimensional vectors, $V^t_i=\{v^t_{i1},\dots,v^t_{in}\}$, where for every $j\in\{1,\dots,n\}$:
\begin{align}
v^t_{ij}
&=\Big[\ \alpha^t_i\big(\tilde{\er}_i(h^1)-\ell(h^1(x^i_j),y^i_j)\big),\ \dots,\ 
  \alpha^t_i\big(\tilde{\er}_i(h^L)-\ell(h^L(x^i_j),y^i_j)\big)\Big].
\end{align}
With this notation, for every $i\in\{1,\dots,k\}$ choosing a random subset 
$S^l_i\subset S^u_i$ corresponds to sampling $m$ vectors from 
$V^t_i$ uniformly without replacement.

For every $i\in\{1,\dots,k\}$, let $U_i=\{u_{i1},\dots,u_{im}\}$ be sampled from $V^t_i$ in that way. 
Then 
\begin{align}
\Phi_t(\bZ)&=F\left(\sum_{i=1}^k\sum_{j=1}^m u_{ij}\right),
\end{align}
where the function $F$ takes as input an $L$-dimensional vector and returns the value of its maximum component.
We now bound $\E_Z\Phi_t(\bZ)$ by applying Lemma~\ref{lem:hoef_convexity} $k$ times:
\begin{align}
\E_Z\Phi_t(\bZ)&=\E_{U_1,\dots,U_k}\ F\Big(\sum_{i=1}^k\sum_{j=1}^m u_{ij}\Big)\\
&=\E_{U_1,\dots,U_{k-1}}\left[\E_{U_k}\Big[F\Big(\sum_{i=1}^{k-1}\sum_{j=1}^m u_{ij}+\sum_{j=1}^m u_{kj}\Big)\Big|U_1,\dots,U_{k-1}\Big]\right]
\intertext{By Lemma~\ref{lem:convexity} $F(x)$ is a convex function. Thus $F(\textit{const}+x)$ is also 
convex and we can apply Lemma~\ref{lem:hoef_convexity} with respect to $U_k$.}
&\leq \E_{U_1,\dots,U_{k-1}}\left[\E_{{\hat U}_k}\Big[F\left(\sum_{i=1}^{k-1}\sum_{j=1}^m u_{ij}+\sum_{j=1}^m{\hat u}_{kj}\right)\Big|U_1,\dots,U_{k-1}\Big]\right]
\intertext{where $\hat U_k=\{u_{ki},\dots,u_{km}\}$ is a set of $m$ vectors sampled from $V^t_k$ \emph{with replacement}.}
&=\E_{U_1,\dots,U_{k-1},{\hat U}_k}\left[F\left(\sum_{i=1}^{k-1}\sum_{j=1}^m u_{ij}+\sum_{j=1}^m{\hat u}_{kj}\right)\right].
\intertext{Repeating the process $k$ times, we obtain}
&\leq \cdots\leq\E_{\hat U_1,\dots,\hat U_{k}}\left[F\left(\sum_{i=1}^k\sum_{j=1}^m {\hat u}_{ij}\right)\right].
\end{align} 
Note that writing the conditioning in the above expressions is just for clarity of presentation, 
since the $U_1,\dots,U_k$ are actually independent of each other.

Switching from the $U$ sets by the $\hat U$ sets in $\Phi$ corresponds to switching from 
random subsets $S^l_i$ to random sets $\tilde{S}_i$ consisting of 
$m$ points sampled from $S^u_i$ uniformly \emph{with} replacement.
Therefore we obtain 
\begin{equation}
\E_Z\Phi_t(Z)=\E_{S^l_1,\dots,S^l_k}\Phi_t(S^l_1,\dots,S^l_k)
            \leq\E_{\tilde{S}_1,\dots,\tilde{S}_k}\Phi_t(\tilde{S}_1,\dots,\tilde{S}_k),
\end{equation}
which allows us to continue analyzing $\E_Z\Phi_t(\bZ)$ in the standard way using Rademacher 
complexities and independent samples. 
%
%
Applying the common symmetrization trick and introducing Rademacher random variables $\sigma_{ij}$ 
we obtain 
\begin{align*}
\Phi_t(\tilde{S}_1,\dots,\tilde{S}_k)
\leq 2
\E_{\sigma}\sup_{h}\sum_{i=1}^k\sum_{j=1}^{m}\sigma_{ij}\alpha^t_i\ell(h(x^i_j),y^i_j).
\end{align*}
We can rewrite this using the fact that $\ell(y,y')=\llbracket y\neq y' \rrbracket=\frac{1-yy'}{2}$:
\begin{align*}
\E_{\sigma}\sup_{h}\sum_{i=1}^k\sum_{j=1}^{m}\sigma_{ij}\alpha^t_i\ell(h(x^i_j),y^i_j)=\E_{\sigma}\sup_{h}\sum_{i=1}^k\sum_{j=1}^{m}\sigma_{ij}\alpha^t_i\frac{1-h(x^i_j)y^i_j}{2}&=\frac{1}{2}\E_{\sigma}\sup_{h}\sum_{i=1}^k\sum_{j=1}^{m}-\sigma_{ij}y^i_j\alpha^t_ih(x^i_j)
\intertext{Since $-\sigma_{ij}y^i_j$ has the same distribution as $\sigma_{ij}$:}
&=\frac{1}{2}\E_{\sigma}\sup_{a(h)\in A}\sum_{i=1}^k\sum_{j=1}^{m}\sigma_{ij}a_{ij}(h),
\end{align*}
where $a_{ij}(h)=\alpha^t_ih(x^i_j)$ and $A=\{a(h):\;h\in \H\}$.
According to Sauer's lemma (Corollary 3.3 in~\cite{MohriBook}):
\begin{equation}
|A|\leq\left(\frac{ekm}{d}\right)^{d}.
\end{equation}
At the same time:
\begin{equation}
\|a\|_2=\sqrt{\sum_{i=1}^k\sum_{j=1}^m(\alpha^t_ih(x^i_j))^2}=\sqrt{m}\sqrt{\sum_{i=1}^k(\alpha^t_i)^2}.
\end{equation}
Therefore, by Massart's lemma (Theorem 3.3 in~\cite{MohriBook}):
\begin{equation}
\E_{\sigma}\sup_{h}\sum_{i=1}^k\sum_{j=1}^{m}\sigma_{ij}\alpha^t_i\ell(h(x^i_j),y^i_j)\leq\frac{1}{2}\sqrt{\sum_{i=1}^k(\alpha^t_i)^2}\cdot\sqrt{2dm\log(ekm/d)}.
\end{equation}
By applying this result for all $t$ we obtain:
\begin{align}
\E_{\bZ}\Phi(\bZ)=\frac{1}{Tm}\sum_{t=1}^T\E_{\bZ}\Phi_t(\bZ)&\leq\frac{1}{Tm}\sum_{t=1}^T\E_{\tilde{S}}\Phi_t(\tilde{S})\leq
\frac{1}{T}\sum_{t=1}^T\sqrt{\sum_{i=1}^k(\alpha^t_i)^2}\cdot\sqrt{\frac{2d\log(ekm/d)}{m}}.
\label{eq:step323}
\end{align}
Combining~\eqref{eq:step322} and~\eqref{eq:step323} with Lemma~\ref{lem:mcdiarmid} we obtain that  for fixed unlabeled samples $S^u_1,\dots,S^u_T$ with probability at least $1-\delta/4$ for all choices of $h_1,\dots,h_T$:
\begin{equation}
\frac{1}{T}\sum_{t=1}^T\tilde{\er}_{\alpha^t}(h_t)\leq\frac{1}{T}\sum_{t=1}^T\eer_{\alpha^t}(h_t)+\frac{1}{T}\|\alpha\|_{2,1}\sqrt{\frac{2d\log(ekm/d)}{m}}
+\frac{1}{T}\|\alpha\|_{1,2}\sqrt{\frac{\log(4/\delta)}{2m}}.
\notag
\end{equation}
By further combining it with~\eqref{eq:step31-ms} we obtain that the following inequality holds uniformly in $h_1,\dots,h_T\in \H$ with probability at least $1-\delta/2$ over 
the sampling of the unlabeled training sets, $S^u_1,\dots,S^u_T$, 
and labeled training sets, $(S^l_i)_{i\in I}$, 
provided that the subset of labeled tasks, $I\subset\{1,\dots,T\}$,
and the task weights, $\alpha^1,\dots,\alpha^T\in\Lambda^I$, 
depend deterministically on the unlabeled training only.
\begin{align}
\notag
\frac{1}{T}\sum_{t=1}^T\er_{\alpha^t}(h_t)\leq\frac{1}{T}\sum_{t=1}^T\eer_{\alpha^t}(h_t)+&\frac{1}{T}\|\alpha\|_{2,1}\sqrt{\frac{2d\log(ekm/d)}{m}}+\frac{1}{T}\|\alpha\|_{1,2}\sqrt{\frac{\log(4/\delta)}{2m}}\\
+&\sqrt{\frac{8(\log T+d\log(enT/d))}{n}}+\sqrt{\frac{2}{n}\log\frac{4}{\delta}}.\!
\label{eq:step3-ms}
\end{align}

The statement of Theorem~\ref{thm:main} follows by combining~\eqref{eq:traingular-ineq-ms} with~\eqref{eq:bound-disc-ms} and~\eqref{eq:step3-ms}.

\end{document}